\documentclass[a4paper,fleqn]{cas-dc} 
\usepackage{color}
\usepackage{stfloats}
\usepackage{natbib}
\usepackage{threeparttable}
\usepackage{bigstrut}
\usepackage[justification=centering]{caption}

\def\tsc#1{\csdef{#1}{\textsc{\lowercase{#1}}\xspace}}
\tsc{WGM}
\tsc{QE}
\tsc{EP}
\tsc{PMS}
\tsc{BEC}
\tsc{DE}

\begin{document}
\let\WriteBookmarks\relax
\def\floatpagepagefraction{1}
\def\textpagefraction{.001}
\shorttitle{Improving the Level of Autism Discrimination through GraphRNN Link Prediction}
\shortauthors{Haonan Sun \textit{et~al.}}

\title [mode = title]{Improving the Level of Autism Discrimination through GraphRNN Link Prediction}                      



\author[1]{Haonan Sun}
\fnmark[1]

\address[1]{College of Medicine and Biological Information Engineering, Northeastern University, 110004 Shenyang, China}
\address[2]{Key Laboratory of Intelligent Computing in Medical Image, Ministry of Education, Northeastern University, 110169 Shenyang, China}
\address[3]{Department of Electrical and Computer Engineering, Stevens Institute of Technology, Hoboken, NJ, USA}

\author[1]{Qiang He}
\fnmark[2]

\author[1]{Shouliang Qi}
\fnmark[3]

\author[3]{Yudong Yao}
\fnmark[4]

\author[1,2]{Yueyang Teng}
\cormark[1]
\fnmark[5]
\ead{tengyy@bmie.neu.edu.cn}


\cortext[cor1]{Corresponding author}
%

\begin{abstract}		
Dataset is the key of deep learning in Autism disease research. However, due to the few quantity and heterogeneity of samples in current dataset, for example ABIDE (Autism Brain Imaging Data Exchange), the recognition research is not effective enough. Previous studies mostly focused on optimizing feature selection methods and data reinforcement to improve accuracy. This paper is based on the latter technique, which learns the edge distribution of real brain network through GraphRNN, and generates the synthetic data which has incentive effect on the discriminant model. The experimental results show that the combination of original and synthetic data greatly improves the discrimination of the neural network. For instance, the most significant effect is the 50-layer ResNet, and the best generation model is GraphRNN,  which improves the accuracy by 32.51$\%$ compared with the model reference experiment without generation data reinforcement. Because the generated data comes from the learned edge connection distribution of Autism patients and typical controls functional connectivity, but it has better effect than the original data, which has constructive significance for further understanding of disease mechanism and development.

\end{abstract}


\maketitle

\section{Introduction}

In the research of brain network science, it is extremely important to select the appropriate dataset for the computer to identify the structural anomalies correctly and efficiently \citep{RN11}. In fact, with the collaborative efforts of researchers, a lot of large datasets have been established, such as Autism Brain Imaging Data Exchange (ABIDE) \citep{RN63,RN64} based on Autism, Alzheimer's Disease Neuroimaging Initiative (ADNI) \citep{RN61} based on Alzheimer's disease, Parkinson's Progression Markers Initiative (PPMI) \citep{RN62} based on Parkinson's disease, etc. Although the data of patients with brain diseases and controls are huge, due to the large data of each individual in brain imaging.In fact, due to difficulties in collection and other reasons, the number of individual samples in the dataset cannot be as sufficient as in other fields. This leads to the direct training is difficult to fully learn the effective information \citep{RN24,RN18,RN23}. Therefore, the prediction model is obviously not universally applicable to individual differences. And it is easy to appear that the training accuracy is high, but it can only get a low accuracy in the test dataset \citep{RN25,RN13}.

It is worth noting that many studies \citep{RN29,RN21,RN19,RN20,RN26,RN22,RN31,RN27,RN30} in ABIDE have not achieved high accuracy in PPMI \citep{RN70,RN69,RN71} and ADNI \citep{RN67,RN66,RN68}. According to the results of previous studies, in addition to the number of samples in the dataset, it may also be due to the significant heterogeneity \citep{RN50,RN48,RN53,RN49,RN51,RN55,RN52} of Autism (which makes it difficult to determine the appropriate classification label standard), the difficulty in distinguishing the Control group from the autistic patients after preprocessing \citep{RN35}, and other factors. How to make the model from the large-scale dataset training, still has high prediction accuracy for the functional connectivity network of patients with brain disease, has been one of the key problems \citep{RN36,RN46,RN13}. 


For this reason, in previous studies, many researchers used manual \citep{RN7,RN60}, Autoencoder (AE) \citep{RN21,RN20} and Sparse Autoencoder (SAE) \citep{RN24,RN25,RN23} to select features. In practical terms, researchers prefer to choose features that improve the dataset, which leads to three problems: 

\begin{enumerate}[i)]
\itemsep=0pt
\item Most of the models only have high accuracy when they are trained on few datasets;
\item It has to identify many features by researchers;
\item The subjectivity of the method is relatively obvious.
\end{enumerate}  

We need to get a method that is generally applicable to all patients and controls brain networks:

Another idea is the augmentation of the dataset \citep{RN73,RN72}. It is worth noting that the traditional image data processing, such as clipping, rotation, scaling, adding noise, blurring, changing the shading or occlusion content, changes the original dataset from the pixel level, but it has little effect on the brain topology graph data. Because graph data focuses more on the structure information of graph, such as the clustering coefficient and local efficiency of nodes, correlation matrix, relevance and efficiency between two nodes, whether the nodes themselves form hub points and the rich-club between them, etc. At the beginning, people used the theory \citep{RN7,RN16} of complex networks to construct or generate graphs with graph theory parameters similar to those in the real world, so that they also have scale-free, small world and other previously proven graph theory properties \citep{RN12,RN60}. However, because the mechanism of brain disease is not clear, the method of thinking that the patients with brain disease have similar graph attributes contains some subjectivity. On the contrary, many normal people have different brain structure attributes, which will cause great interference to our model of generating brain network.

Fortunately, in recent years, with the maturity of image data generation model \citep{RN36,RN34,RN39}, many excellent models for graph data generation have been built analogically, such as GraphVAE \citep{RN8}, VGAE \citep{RN9}, GraphGAN \citep{RN45} and GraphRNN \citep{RN32}, etc. They have achieved good results on small graph datasets, such as proteins, citation networks, drug molecules, etc. Unfortunately, the existing discrimination methods are difficult to distinguish the differences between the generated graph and the real graph effectively and quantitatively \citep{RN32}. Many of them rely on the simple comparison of visual observation or statistical data \citep{RN8,RN45} (this is indeed feasible in the field of drug molecules and other small graphs).

Note that the existing graph generation models, such as GraphGAN \citep{RN45} is implemented with Softmax when the generator estimates the connection distribution. Softmax function treats each node fairly and ignores the topological structure of brain network when it completes normalization. In GraphGAN, in order to add network information, the author proposes Graph Softmax \citep{RN45} to calculate the estimated connection distribution, but the Softmax function with structure information needs to satisfy three conditions: In short, we need to satisfy a specific probability distribution, the connectivity probability of two vertices on the graph should decrease with the increase of the shortest distance, and the calculation of connectivity distribution should only cover a small number of nodes (such as the nodes close to ${v_c}$). Obviously, these conditions are not suitable for the study of brain networks. In addition, the limitation of GraphVAE is that the amount of computation is much larger than GraphRNN, and it is difficult to compute the graph with more nodes, such as brain network graph \citep{RN32}. We note that the original intention of RNN is to simulate the working mechanism and training process of some neurons in the human brain \citep{RN58,RN57,RN54}. Although we can't build as many neurons as the human brain, we hope to achieve better results for specific tasks. In recent years, studies on the mechanism of brain dynamics have proved that RNN can be used to simulate the human brain \citep{RN37}. 

So a natural idea is whether these models can be applied in the field of brain diseases, and whether the data generated is enough to replace the real-world brain graph \citep{RN36,RN41,RN47}. Brain formation is also a new topic in recent years. Its ultimate goal and prospect are very promising. One of them is to simulate human brain development and predict brain diseases \citep{RN7,RN36,RN1,RN35,RN16}. In recent years, Deep reinforcement learning (RL) has developed in depth and breadth in the mechanism of single brain mechanism \citep{RN10}. Moreover, due to the strong specificity of brain diseases such as Autism and many subtypes of patients, we hope to find a general method to eliminate a few differences among patients and train them to learn their common model.

We get inspiration from the generative teaching networks (GTNs) proposed by \citet{RN33}. This model generates completely artificial data to guide the model to have faster efficiency and learning speed. The model generates completely artificial graph data to guide the model to have faster learning speed. In order to further improve the accuracy of the model, we use GraphRNN to learn the edge connection distribution of patients and controls in the data generation step, and generate incentive data for the two types of data, so that the discrimination model has higher efficiency and accuracy.

So our final idea is to learn the distribution of brain networks in the real world based on GraphRNN, and then use it to generate incentive data to improve the identification ability of the original deep learning model.
 

\section{Methodology}

The existing datasets are preprocessed with MATLAB, and the optimal parameters such as atlas, threshold and preprocessing method are selected to make the subsequent experiments unaffected. Using the new dataset after processing, GraphRNN is allowed to learn the data of Autism and Control group respectively, and generate the dataset containing false Autism and Control. At the same time, the discrimination model used after training also selects the optimal parameters and model, and the model can be used to judge the quality of the previously generated dataset. Finally, the real brain network and the generated dataset are combined and disrupted, and the optimal discriminant model is trained to predict the real binary brain network dataset. Because it comes from the original dataset, it is essentially to extract the features of the ground truth data, and finally improve the existing model.

\subsection{Data preprocessing}

This paper refers to the work of \citet{RN28} to solve how to extract times series to build a functional connectome. Because ABIDE(\url{http://fcon_1000.projects.nitrc.org/indi/abide/}) contains data from Caltech, CMU, KKI and other 17 Organizations. In addition to the parameters we can control, such as TR, TE, Flip angle, Slices, there are also uncontrollable factors, such as Voxel size, Slice thickness, Participant instructions (e.g. eyes open vs. closed), Recruitment strategies (age-group, IQ-range, level of impairment, treatment history and acceptable comorbidities). This leads to the problem of heterogeneous graph which is difficult to avoid.

In previous studies\citep{RN29,RN21,RN20,RN22,RN24,RN31,RN18,RN23,RN30}, the brain graph data is usually the functional connectivity matrix between the set of brain region of interest (ROI). The data is based on the expert's examination results, excluding most of the largely incomplete brain coverage, high movement peaks, ghosting and other scanner artifacts. At the same time, preprocessing without deleting data is also carried out, which includes slice-timing correction, image realignment to correct for motion, and intensity normalization\citep{RN28,RN11}. We keep some datasets which are still controversial in preprocessing methods for later discussion. In fact, through expert inspection data, these datasets have been smaller than the original dataset of 1112 subjects in varying degrees.

We used the same preprocessing pipeline\citep{RN40} to ensure consistency with existing studies\citep{RN28,RN31,RN27}, which involves motion correction, slice timing correction, skull striping and nuisance signal regression. Whether global mean intensity normalization and band-pass filtering (0.01-0.1Hz) is processed is currently controversial, so we will carry out experimental verification.

\subsection{Brain network and problem definition}

\begin{figure}
	\centering
	\includegraphics[width = 0.95\linewidth]{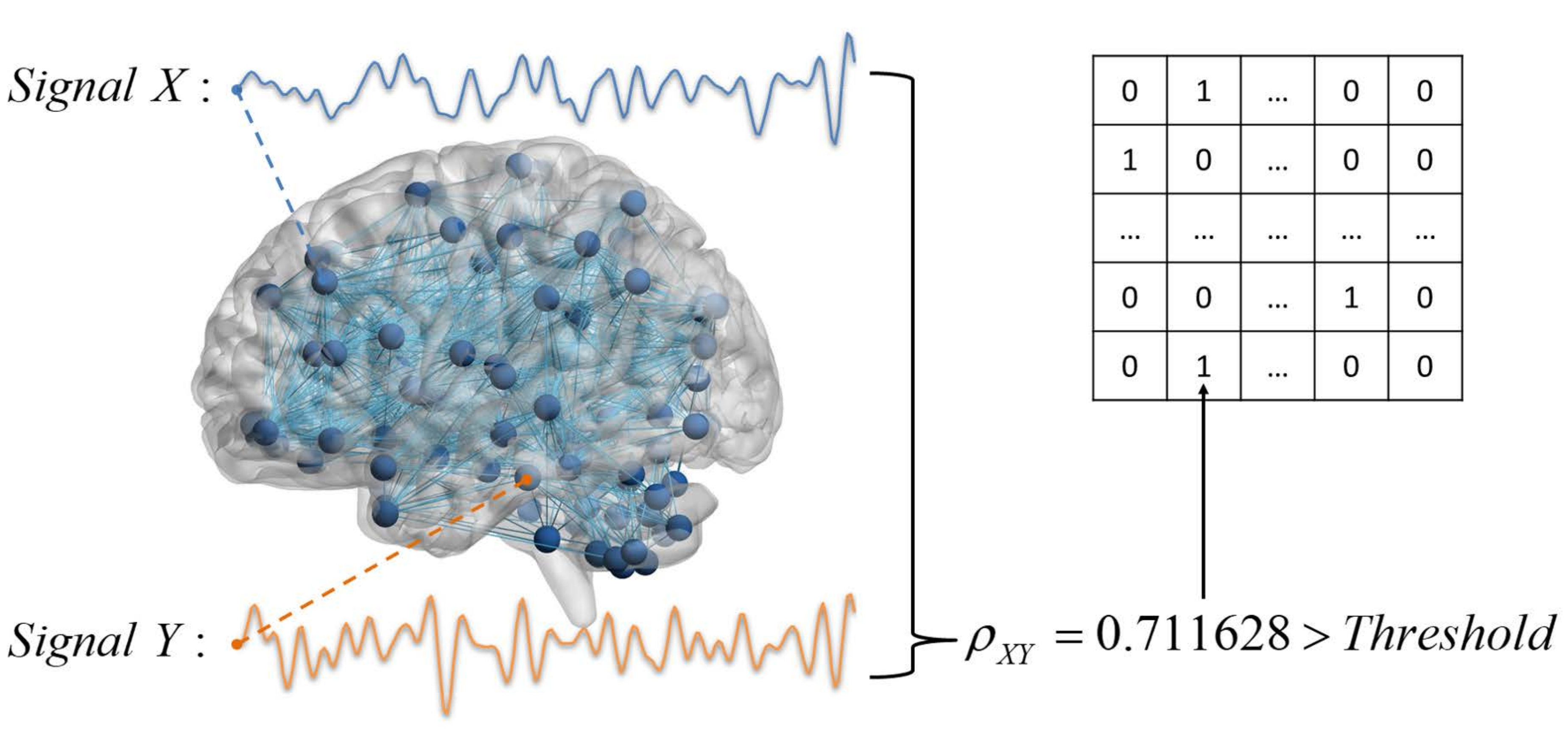}
	\caption{The edge connection of brain nodes is transformed into adjacency matrix under a specific threshold parameter.}
	\label{fig:1}
\end{figure}

In this paper, regions of interests (ROIs) under different atlases are defined as network nodes\citep{RN42,RN59}, so that the corresponding graph data of the same brain under different atlases are also different, which provides more reference groups for the training and verification of the later model. At the same time, the edge of brain connection is crucial to comprehend the deep structure of graph and explain the similarity between eigenvectors\citep{RN44}. Then, for the time series of any two network nodes $X$ and $Y$, the Pearson correlation coefficient matrix (\textcolor{black!80}{Formula.~\protect\ref{eq1}}) between them can be obtained. In the experiment, we add a threshold to control the sparsity of the network\citep{RN60,RN14}. If the threshold exceeds, it means that there is edge connection between the two nodes, otherwise, there is no edge connection. We can use the binary adjacency matrix to represent the relationship between a patient or a control brain network node (\textcolor{black!80}{Fig.~\protect\ref{fig:1}}), and transform the resting-state fMRI time series data into the ground truth data needed for the later generation of the graph.

\begin{eqnarray}\label{eq1}
	\rho_{MN} = \frac{{\sum\limits_{i = 1}^n {(M_i - \bar M)(N_i - \bar N)} }}{{\sqrt {\sum\limits_{i = 1}^n {{{(M_i - \bar M)}^2}} } \sqrt {\sum\limits_{i = 1}^n {{{(N_i - \bar N)}^2}} } }}
\end{eqnarray}

After preprocessing, the brain network can be regarded as an undirected graph $G=(V,E)$, which is defined by the node set $V=\left\{ {{v_1},{v_2}, \ldots ,{v_n}} \right\}$ representing the ROIs and the definition of edge set $E=\left\{ {({v_i},{v_j})|{v_i},{v_j} \in V} \right\}$ that represents the association between ROIs.Previously, we obtained its adjacency matrix by preprocessing. In fact, because the representation of adjacency matrix is not unique, we assume that the nodes selected under the specified brain region template have a sort  , and map the nodes to the adjacency matrix. More precisely, $\left\{{\psi ({v_1}),\psi ({v_2}), \ldots ,\psi ({v_n})}\right\}$ is the permutation of node set $\left\{{{v_1},{v_2}, \ldots ,{v_n}}\right\}$.

We define $\Psi$ the possible arrangement set of all brain network nodes (if the template has $n$ brain regions, then $\Psi$ has $n!$ arrangements). In the last part we get its adjacency matrix by preprocessing. Actually, for the same brain network graph, the adjacency matrix representation is not unique. Therefore, we need to determine in advance a set of appropriate ranking $\psi$ of the selected nodes under the brain regions’ mask, which is suitable for all patients and typical controls.  

\begin{figure}
	\centering
	\includegraphics[width = 0.95\linewidth]{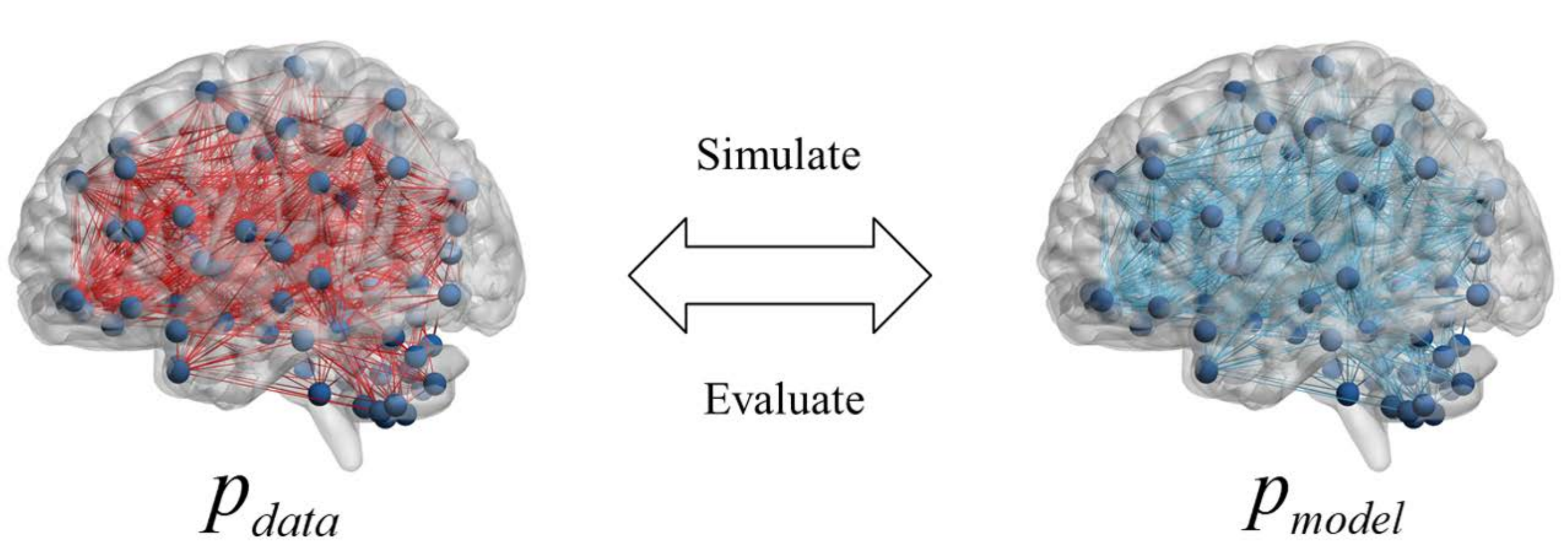}
	\caption{The purpose of training is to make the distribution of the model approximate to that of the real world brain network.}
	\label{fig:2}
\end{figure}

Then for the node arrangement under a group of brain area templates, the brain network $G$ of patients or control group can be represented as a group ${A^\psi } \in {\mathbb{R}^{n \times n}}$, where $A_{i,j}^\psi \in \mathbb{1}\left[ {(\psi ({v_i}),\psi ({v_j})) \in E} \right]$. Note that in the adjacency matrix set, the elements of ${A^\psi } = \left\{ {\left. {{A^\psi }} \right|\psi  \in \Psi } \right\}$ correspond to the same underlying brain network graph. Our goal of generating the model is to learn the distribution ${p_{model}}(G)$ of the existing data brain network from the brain observation graph data $\mathbb{G} = \left\{ {{G_1},{G_2}, \ldots ,{G_s}} \right\}$ sampled from the real world $p(G)$(\textcolor{black!80}{Fig.~\protect\ref{fig:2}}).

We assume that the order $\psi$ of each brain node is equal probability, i.e., $p(\psi ) = \frac{1}{{n!}},\forall \psi  \in \Psi$. Therefore, we want to use the existing brain graphs for augmentation. Because the traditional pixel level augmentation methods are not suitable for graphs, in fact, the generated brain graphs also have exponential representation, which is different from the previous generation models for images, texts and time series. In addition, the traditional method is generally single input training image, but the graph field can integrate multiple graphs into a large graph, and then use annotation to locate the features or labels of subgraphs.   

\subsection{GraphRNN: Deep generative models for brain network}

The key idea of GraphRNN \citep{RN32} is to represent graphs under different node orderings as sequences, and then to build an autoregressive generative model on these sequences. The advantage of this method is that it allows us to model large graphs with complex edge dependencies (such as the brain functional connectivity graph), which is not affected by the common shortcomings of other generation methods (such as GraphGAN \citep{RN45} and GraphVAE \citep{RN8}). And GraphRNN can introduce the idea of breadth first search (BFS) node traversal to greatly reduce the complexity of all possible node sequence learning. In this autoregressive framework, the complexity of the model can be greatly reduced by sharing weights with recurrent neural networks (RNNs).

\begin{figure*}
	\centering
	\includegraphics[width=0.95\linewidth]{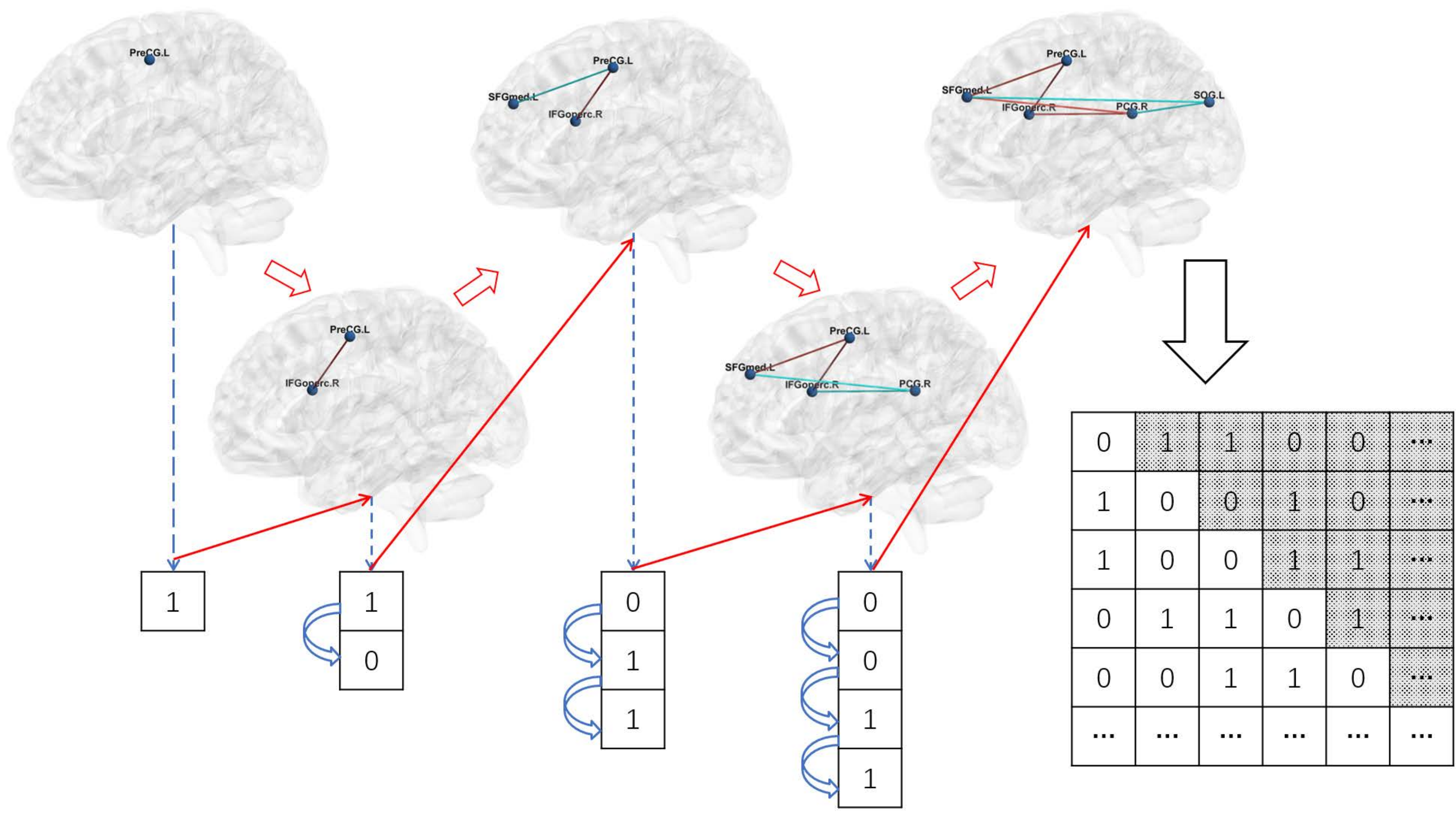}
	\caption{The process of brain network deduction by GraphRNN. The red arrows denote the graph level RNN that encodes the graph state vector ${h_i}$ in its hidden state updated by the predicted adjacency vector $S_i^\psi$ for $\psi ({v_1})$. The blue arrows represent the edge level RNN whose hidden state is initialized by the graph level RNN used to predict the adjacency vector $S_i^\psi$ for $\psi ({v_1})$.}
	\label{fig:3}
\end{figure*}

\textcolor{black!80}{Fig.~\protect\ref{fig:3}} shows the process of GraphRNN learning brain network step by step. The main idea is that we decompose graph generation into a process of generating a series of nodes (through graph level RNN), and the other process generates a series of edges (through edge level RNN) for each newly added node.
Note that this is only a traversal method for learning the ground truth graph of the known real world. But for our trained model to generate new brain network nodes, it will help us to explore the mechanism of Autism, because most Autism is formed during the early brain development.

Graph is a complex data structure. In order to optimize graph operation, we first define the mapping ${f_S}$ from graph to a sequence. For a graph $G \sim p(G)$ with $n$ brain nodes in order $\psi$, we have 
\begin{eqnarray}\label{eq2}
{S^\psi } = {f_S}(G,\psi ) = (S_1^\psi ,S_2^\psi , \ldots ,S_n^\psi )
\end{eqnarray}
where each element $S_i^\psi  \in {\{ 0,1\} ^{i - 1}},i \in \left\{ {1,2, \ldots ,n} \right\}$ in the sequence is an adjacency vector representing the edges between brain network node $\psi ({v_i})$ and the previous brain network nodes $\psi ({v_j}),j \in \left\{ {1,2, \ldots ,i - 1} \right\}$:
\begin{eqnarray}\label{eq3}
S_i^\psi  = {(A_{1,i}^\psi ,A_{2,i}^\psi , \ldots ,A_{i - 1,i}^\psi )^T},\forall i \in \left\{ {2,3, \ldots ,n} \right\}
\end{eqnarray}
We will not consider the self-connection within a certain brain region, so we prohibit self-loops and initialize $S_1^\psi$ as an empty vector. For undirected brain graphs, ${S^\psi }$ determines a unique graph $G$. In this way, we transform the learning $p(G)$ which is difficult to define in the sample space into the observation value of ${S^\psi }$ under the brain region mask sampling. Since ${S^\psi }$ is a sequence, it can be modeled. Then the $p(G)$ of the whole graph can be transformed into:
\begin{eqnarray}\label{eq4}
	p(G) = \sum\limits_{{S^\psi }} {p({S^\psi })}
\end{eqnarray}
Where $p({S^\psi })$ is the distribution that we want to learn from the generative model. Due to the sequence property of ${S^\psi }$, we can further decompose $p({S^\psi })$ into the product of conditional distributions on each brain node:
\begin{eqnarray}\label{eq5}
p({S^\psi }) = \prod\limits_{i = 1}^{n + 1} {p(\left. {S_i^\psi } \right|S_1^\psi , \ldots ,S_{i - 1}^\psi )}
\end{eqnarray}
We will set $S_{n + 1}^\psi $ as end of sequence, and using $p(\left. {S_i^\psi } \right|S_{ < i}^\psi )$ to simplify $p(\left. {S_i^\psi } \right|S_1^\psi , \ldots ,S_{i - 1}^\psi )$.


It is observed that $p(\left. {S_i^\psi } \right|S_{ < i}^\psi )$ has to capture how the current node $\psi ({v_i})$ links to the previous node based on the connection among the previous nodes, so RNN can be used to parameterize the distribution of $p(\left. {S_i^\psi } \right|S_{ < i}^\psi )$.  In order to achieve a scalable model, the neural network can share the weight in all time steps $i$. The state transition function and output function of RNN can be expressed as:
\begin{eqnarray}\label{eq6}
\begin{split}
	{h_i} &= {f_{trans}}({h_{i - 1}},S_i^\psi )    \\
	{\theta _{i + 1}} &= {f_{out}}({h_i})
\end{split}
\end{eqnarray}
Where ${h_i} \in {\mathbb{R}^{\text{d}}}$ is a vector encoding the state of the graph so far, $S_i^\psi$ is the adjacency vector of the most recently generated node $i$, and ${\theta _{i + 1}}$ specifies the adjacency vector distribution of the next node. Where $S_i^\psi$ follows the arbitrary distribution ${P_{{\theta _{i + 1}}}}$ of binary vectors.

The loss function of each time is defined as the binary cross entropy loss. Then the RNN model minimizes the following loss function:
\begin{eqnarray}\label{eq7}
	L =  - \frac{1}{{N - 1}}\sum\limits_{i = 0}^{N - 1} {[y_i^*\log ({y_i}) + (1 - y_i^*)\log (1 - {y_i})]}
\end{eqnarray}
Where $y_i^*$ is the corresponding value of binary matrix obtained by preprocessing, ${y_i}$ is the output of RNN unit in each step, and $N$ is the number of all possible connections, which is related to the number of nodes selected for brain mask.
 
\section{Experiments}

\subsection{Datasets description and effect evaluation}

The data support of this paper comes from the ABIDE dataset \citep{RN64} of the preprocessed connectors project (PCP) \citep{RN63}, which is a public dataset of neuroimaging data from 539 patients and 573 controls. The whole dataset is provided by 16 institutions around the world \citep{RN28}. In this experiment, resting state functional MRI (rs-fMRI) data was selected. In order to facilitate comparison with other studies, we selected the data processed by C-PAC (The Configurable Pipeline for the Analysis of Connectomes) \citep{RN40}. According to the research process of \citet{RN28}, we only keep the images whose average frame by frame displacement of the function image is less than 0.2, and the rest of the data that may have negative impact on the model results have been eliminated in the preprocessing.

For the part of data impact assessment, t-distributed stochastic neighbor embedding (t-SNE) is used to evaluate the dataset qualitatively and visually, and residual neural network (ResNet) is used as the classifier of the whole model for quantitative comparison. For other models, we use real-world brain datasets for pre-experimental classification, and reproduce the effect of graph neural network proposed in recent years. Unfortunately, it does not achieve better accuracy than ResNet. In order to control the influence of other factors and stabilize the model, the classifier is fixed.

There are two purposes for ResNet to quantify the classification effect by accuracy: One is to determine which preprocessing method is effective for model learning in the initial part of model training, so as to determine an appropriate preprocessing method and provide a more effective ground truth graph for generating model; the other is to provide a method to measure the quality of the generated graphs. Because the traditional measurement methods (histogram, PSNR, SSIM, NMI) are difficult to measure the quality of brain topology graph, and the distance index in graph theory is also difficult to measure the deep features. So we expect to use deep layer ResNet to detect the difference of the deep features between autistic patients and typical controls.

\subsection{Defining nodes based on brain atlas}

In the previous section, we use \textcolor{black!80}{Formula.~\protect\ref{eq1}} to constructed the rules of the edge of brain network.  The rules in this section are to make rules and experiments for nodes. Because the principle of using different atlas is different, we hope to find the impact of these atlases on the model. The four brain atlas regions selected in this paper are also collected through C-PAC:
\begin{enumerate}[(A)]
	\item Automated Anatomical Labeling (AAL): The AAL atlas is based on Montreal Neurological Institute (MNI) regions drawn from T1 weighted images of a single subject. The main method is to segment the anatomical region manually, and then calibrate the program automatically. In this paper, we select the mask which contains 116 brain region nodes.
	\item Harvard-Oxford (HO): The HO atlas is a probability atlas. The T1 weighted image is affine transformed to MNI space by FSL, and then the transformation is applied to a single label. It was created by the FMRIB analysis team in Oxford, UK. In this paper, we select the mask which contains 111 brain region nodes.
	\item Craddock 200 (CC200): The CC200 atlas refers to the research of Craddock et al. The principle of brain atlas formation through grey matter master is to use normalized cut and spectral clustering algorithm, and the final brain atlas has 200 brain region nodes.
	\item Craddock 400 (CC400): This brain atlas is an expanded version of cc200 with 400 brain region nodes.
\end{enumerate}
\begin{figure}
	\centering
	\includegraphics[scale=.2]{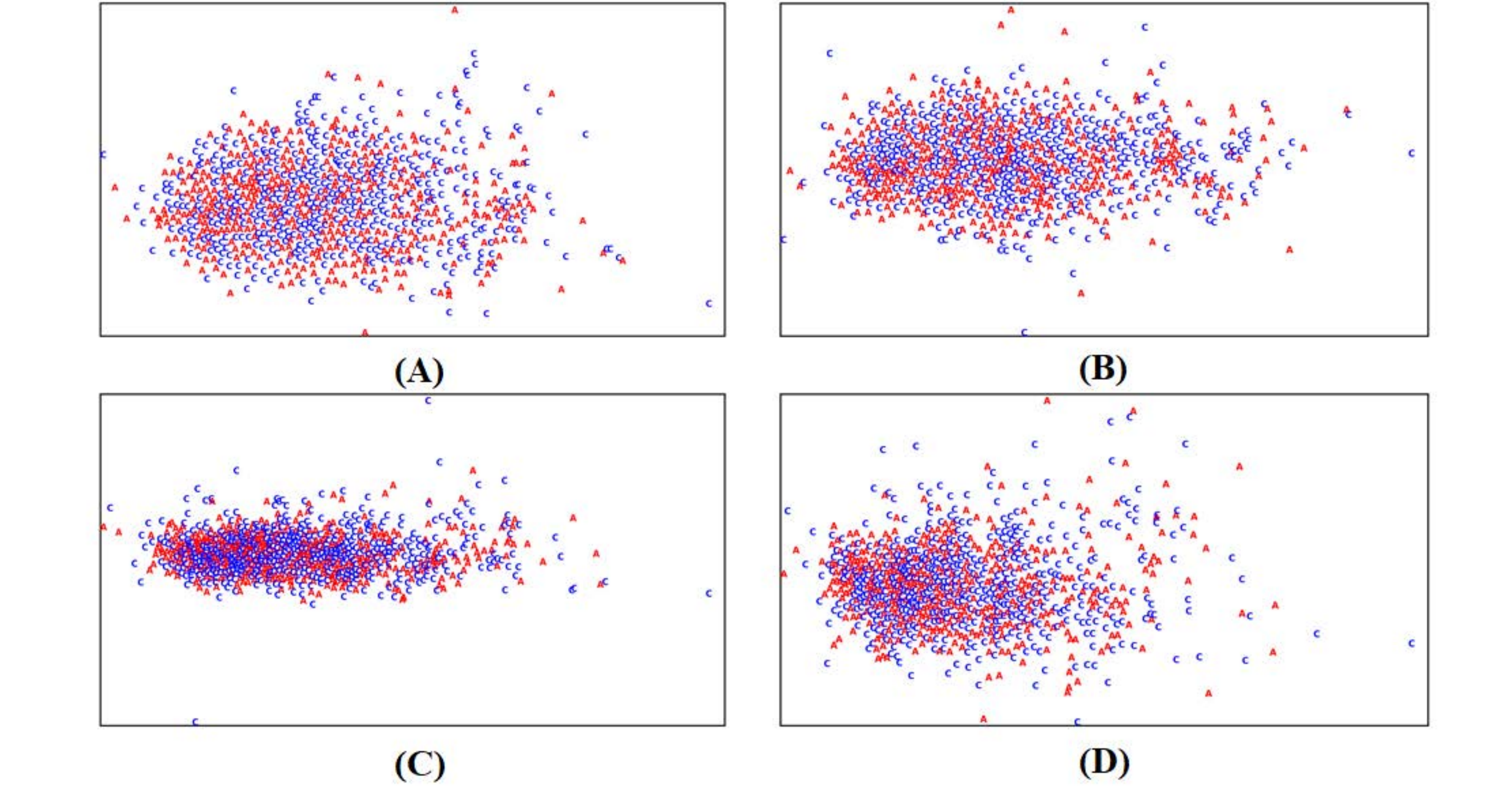}
	\caption{t-SNE is used to embed ABIDE dataset under different masks (A)AAL, (B)HO, (C)CC200, (D)CC400 into two-dimensional space for visualization, in which "A" represents autistic patients and "C" represents typical controls.}
	\label{fig:4}
\end{figure}
The graph generated by different atlas for the same subject is heterogeneous, because the node structure and edge connection are different. In order to preliminarily estimate the advantages and disadvantages of these datasets, this paper uses t-distributed stochastic neighbor embedding (t-SNE) to reduce the dimension of the data set before the experiment, and then makes a qualitative analysis of the mask selection.

Due to the indeterminacy of data sources in band-pass filtering and global signal regression \citep{RN63}, in addition to the number of brain nodes directly affected by the template. Under the same threshold conditions, we conducted ablation experiments to determine whether the real data need filtering and global signal regression, so as to determine the effective preprocessing data source and generate data for the later model. We will carry out model learning (\textcolor{black!80}{Table~\protect\ref{tbl1}}) according to the four kinds of data generated by the above preprocessing, and select the appropriate preprocessing strategy through the accuracy.

\begin{table}[width=.95\columnwidth,cols=4,pos=h]
	\begin{threeparttable}
		\centering
		\caption{The four kinds of data generated by the above preprocessing.}
		\begin{tabular*}{\tblwidth}{@{} CCC@{} }
			\toprule[2pt]
			\textbf{Method} & \textbf{Mask} & \textbf{Accuracy(\%)} \\
			\midrule
			\multirow{4}[2]{*}{\textbf{Filter+Global}} & \textbf{AAL116} & \textbf{61.80} \\
			& \textbf{HO111} & \textbf{55.06} \\
			& \textbf{CC200} & \textbf{52.87} \\
			& \textbf{CC400} & \textbf{53.93} \\
			\midrule
			\multirow{4}[2]{*}{\textbf{Filter}} & \textbf{AAL116} & \textbf{68.53} \\
			& \textbf{HO111} & \textbf{65.52} \\
			& \textbf{CC200} & \textbf{59.55} \\
			& \textbf{CC400} & \textbf{61.63} \\
			\midrule
			\multirow{4}[2]{*}{\textbf{Global}} & \textbf{AAL116} & \textbf{59.55} \\
			& \textbf{HO111} & \textbf{59.55} \\
			& \textbf{CC200} & \textbf{52.81} \\
			& \textbf{CC400} & \textbf{57.30} \\
			\midrule
			\multirow{4}[2]{*}{\textbf{/}} & \textbf{AAL116} & \textbf{59.55} \\
			& \textbf{HO111} & \textbf{55.68} \\
			& \textbf{CC200} & \textbf{51.69} \\
			& \textbf{CC400} & \textbf{57.30} \\
			\bottomrule[2pt]
		\end{tabular*}
		\label{tbl1}
		\begin{tablenotes}
			\footnotesize
			\item[*] \textbf{For the preprocessing method including global signal correction, the global mean signal includes interference variable regression. After variable regression, band-pass filter (0.01$\sim$0.1 Hz) was used.}
		\end{tablenotes}
	\end{threeparttable}
\end{table}

\textcolor{black!80}{Fig.~\protect\ref{fig:4}} reveals that ABIDE dataset is difficult to classify under each mask, which is consistent with the fact that most studies have low accuracy. Therefore, we optimize the dataset and use convolutional neural network to quantify the effect of the dataset. Since the accuracy of the model is relatively stable when filtering without global signal regression and using the AAL brain atlas with 116 nodes (\textcolor{black!80}{Table~\protect\ref{tbl1}}), the above strategy is adopted in the later verification. But so far, we only select the appropriate dataset, and do not operate on the data, so we carry out further evaluation experiments to generate data learning and reinforce the original model, and then use ResNet to quantify the effect of the dataset.

\subsection{Ablation experiment of model parameters}

First of all, we conducted threshold selection experiments for single source real data. Based on the threshold of most previous research experiments, we took a group of larger values, a group of smaller values, a group of moderate values, and adaptive threshold based on Otsu method (the threshold determined by the maximum inter class variance) to compare with each other. The 50-layer ResNet model was used in the experiment. All the data were evaluated by AAL mask with 116 brain nodes. The number of edges depends on the threshold. The maximum number of edges is the original matrix.

\begin{figure*}
	\centering
	\includegraphics[width=0.95\linewidth]{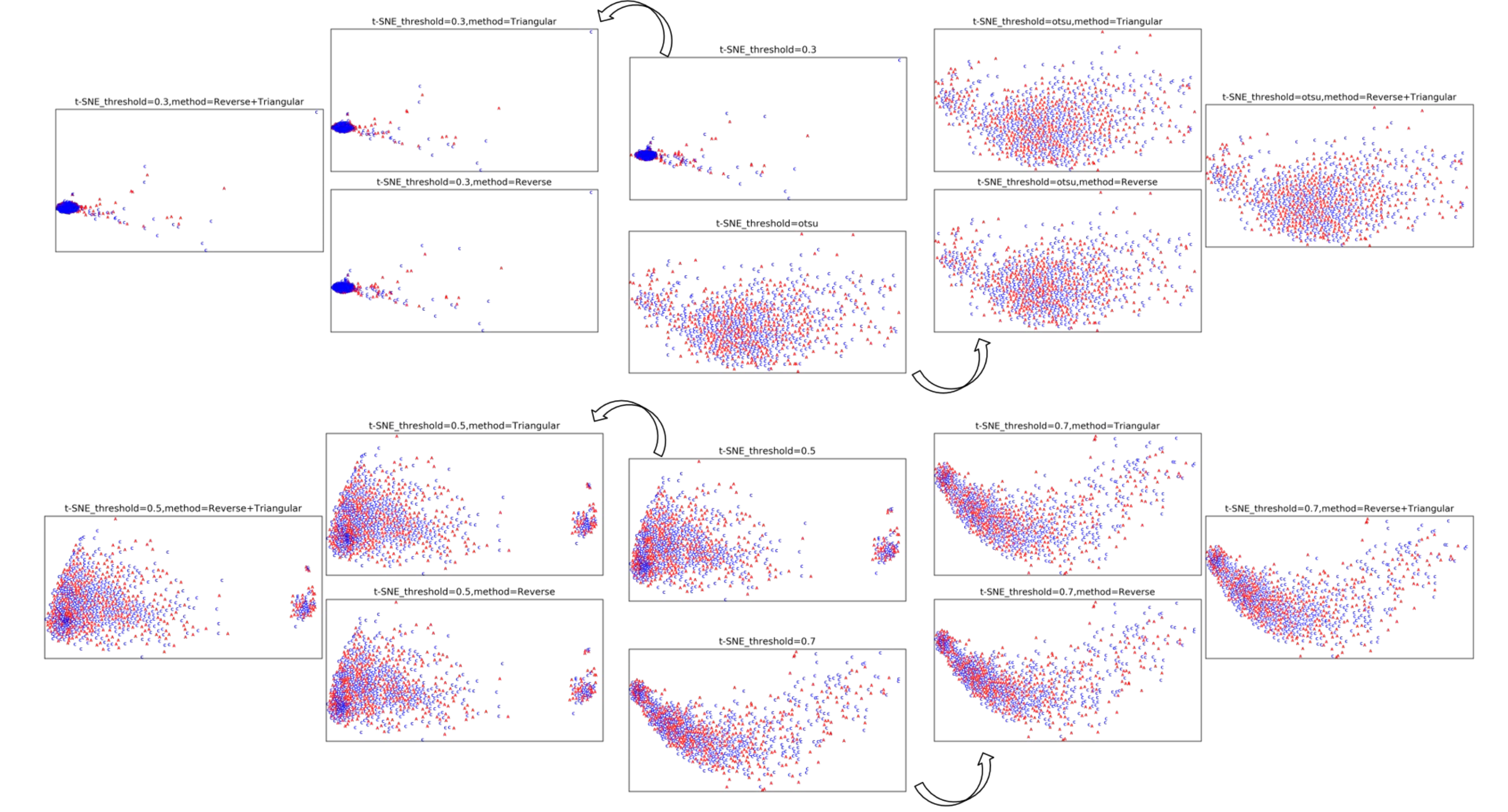}
	\caption{Using t-SNE to embed datasets of different thresholds and methods into two-dimensional space for qualitative comparison, other preprocessing methods almost have no effect on datasets, while the impact of threshold is very considerable.}
	\label{fig:5}
\end{figure*}

In order to verify whether the common reversing and triangular matrix in many studies have an impact on the model, we also reversed the brain connectivity network dataset under different threshold processing (that is, the edge connection is set to "0", the connection without edge is set to "1") and preserved the upper triangular matrix. Before the quantitative comparison experiment, we compared the datasets visually and qualitatively (\textcolor{black!80}{Fig.~\protect\ref{fig:5}}).
Finally, we put the original coefficient matrix into the model for comparison. The matrix is essentially a weight graph, and the edge (without considering the brain region self-connection) is formed between each node, and the weight of the edge represents the correlation coefficient of the average time series of the two brain regions.\textcolor{black!80}{Table~\protect\ref{tbl2}} shows the discrimination effect of these datasets under ResNet50.

Because there are too many edges, we choose the threshold=0.5 to reduce the number of edges, but the accuracy will not be significantly affected.

\begin{table*}[htbp]
	\begin{threeparttable}
		\centering \caption{The performance experiment of the ResNet50 model under the same atlas, different thresholds and preprocessing.}
	  	\centering
	  	\begin{tabular}{c|c|c|c|c}
	  	\toprule[2pt]
	  	\textbf{Threshold} & \textbf{Other preprocessing} & \textbf{Accuracy(\%)} & \textbf{Number of Node} & \textbf{Number of Edge} \\
	  	\midrule
	  	\multirow{4}[2]{*}{\textbf{0.3}} & \textbf{$/$} & \textbf{52.28} &       &  \\
	  	& \textbf{Reverse} & \textbf{56.81} & \textbf{A=46748} & \textbf{A=1135985} \\
	  	& \textbf{Triangular} & \textbf{53.71} & \textbf{C=54288} & \textbf{C=1366380} \\
	  	& \textbf{R$+$T} & \textbf{53.45} &       &  \\
	  	\hline
	  	\multirow{4}[2]{*}{\textbf{0.5}} & \textbf{$/$} & \textbf{60.80} &       &  \\
	  	& \textbf{Reverse} & \textbf{58.27} & \textbf{A=46748} & \textbf{A=432649} \\
	  	& \textbf{Triangular} & \textbf{57.14} & \textbf{C=54288} & \textbf{C=535424} \\
	  	& \textbf{R$+$T} & \textbf{59.43} &       & \\
	  	\hline
	  	\multirow{4}[2]{*}{\textbf{0.7}} & \textbf{$/$} & \textbf{52.13} &       &  \\
	  	& \textbf{Reverse} & \textbf{52.21} & \textbf{A=46748} & \textbf{A=74421} \\
	  	& \textbf{Triangular} & \textbf{57.95} & \textbf{C=54288} & \textbf{C=96445} \\
	  	& \textbf{R$+$T} & \textbf{54.52} &       &  \\
	  	\hline
	  	\multicolumn{1}{c|}{\multirow{4}[2]{*}{\textbf{Otsu threshold}}} & \textbf{$/$} & \textbf{54.41} &       &  \\
	  	& \textbf{Reverse} & \textbf{58.09} & \textbf{A=46748} & \textbf{A=1527691} \\
	  	& \textbf{Triangular} & \textbf{57.95} & \textbf{C=54288} & \textbf{C=1804461} \\
	  	& \textbf{R$+$T} & \textbf{56.74} &       &  \\
	  	\hline
	  	\multicolumn{1}{c|}{\multirow{4}[2]{*}{\textbf{Original}}} & \multicolumn{1}{c|}{\multirow{4}[2]{*}{\textbf{$/$}}} & \multirow{4}[2]{*}{\textbf{61.36}} &       & \\
	  	&       &       & \textbf{A=46748} & \textbf{A=2688010} \\
	  	&       &       & \textbf{C=54288} & \textbf{C=3121560} \\
	  	&       &       &       & \\
	  	\bottomrule[2pt]
	  	\end{tabular}
		\label{tbl2}
		\begin{tablenotes}
			\footnotesize
			\item[*] \textbf{A=Autism, C=Control, Reverse means to swap 0 and 1 in the graph, Triangular denotes Preserving the triangular matrix, Original uses the unprocessed Pearson correlation matrix.}
			\item[**] \textbf{Note that we use the same model and model parameters here.}
		\end{tablenotes}
	\end{threeparttable}
\end{table*}

Considering the influence of classification model, we compare the influence of various ResNet and VGG on classification effect. In this way, we determine the ResNet discriminant model which can not only provide more effective ground truth graph for the generated model, but also measure the quality of the generated graphs.
\begin{figure}
	\centering
	\includegraphics[width = 0.95\linewidth]{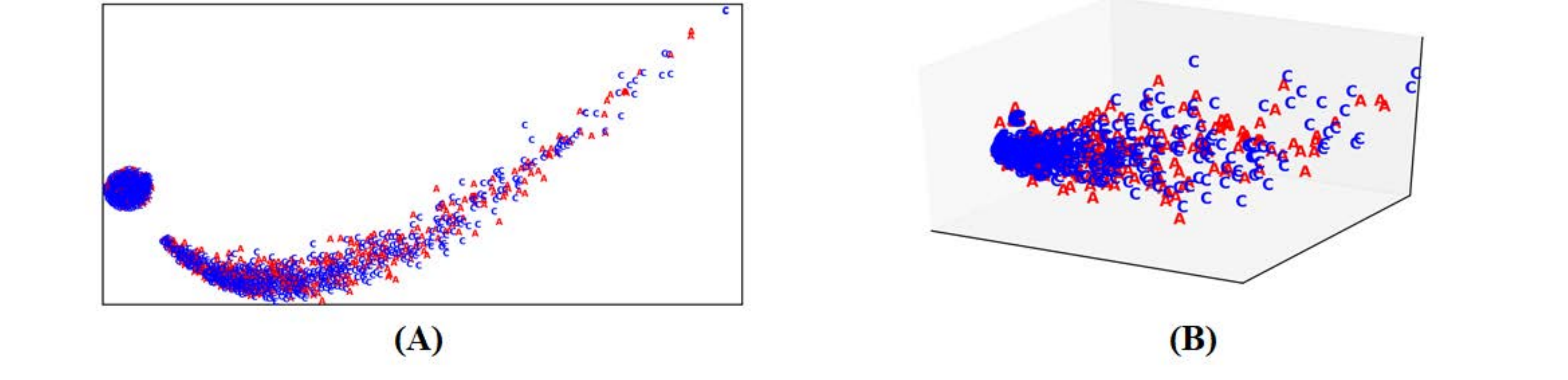}
	\caption{T-SNE is used to embed the generated dataset into (A)two-dimensional and (B)three-dimensional space for qualitative comparison.}
	\label{fig:6}
\end{figure}
The specific step is to train and adjust the parameters of the original threshold dataset, and take the optimal discrimination ability of the model as the benchmark. Then, the generated model generates a nearly equal volume of Autism dataset and Control dataset (\textcolor{black!80}{Fig.~\protect\ref{fig:6}} is the visualization of the dataset). After eliminating the obvious failure and sparse data, the total data set capacity is $871\pm10$.It does not change other parameters, and only uses the generated data to train the original different models. As an experiment to select discriminant model, the original brain network dataset is used to verify the effect of the trained model (\textcolor{black!80}{Table~\protect\ref{tbl3}}).
\begin{table*}[htbp]
	\begin{threeparttable}
		\centering
		\caption{The performance of raw data, generated data and Mixed data(raw + generated) training model to forecast raw data in different models.}
		\centering
		\begin{tabular}{c|c|c}
			\toprule[2pt]
			\textbf{Data used} & \textbf{Model} & \textbf{Accuracy(\%)} \\
			\midrule
			\multirow{5}[2]{*}{\textbf{Raw data training model to forecast raw data (threshold=0.5)}} & \textbf{ResNet-18} & \textbf{54.66} \\
			& \textbf{ResNet-34} & \textbf{58.30} \\
			& \textbf{ResNet-50} & \textbf{60.80} \\
			& \textbf{ResNet-101} & \textbf{52.61} \\
			& \textbf{VGG16} & \textbf{52.27} \\
			\midrule
			\multirow{5}[2]{*}{\textbf{Generated data training model to forecast raw data (threshold=0.5)}} & \textbf{ResNet-18} & \textbf{71.93} \\
			& \textbf{ResNet-34} & \textbf{74.09} \\
			& \textbf{ResNet-50} & \textbf{76.48} \\
			& \textbf{ResNet-101} & \textbf{71.25} \\
			& \textbf{VGG16} & \textbf{51.71} \\
			\midrule
			\multirow{5}[2]{*}{\textbf{Mixed data training model to forecast raw data(threshold=0.5)}} & \textbf{ResNet-18} & \textbf{58.70} \\
			& \textbf{ResNet-34} & \textbf{63.04} \\
			& \textbf{ResNet-50} & \textbf{69.57} \\
			& \textbf{ResNet-101} &\textbf{57.25} \\
			& \textbf{VGG16} & \textbf{48.86} \\
			\bottomrule[2pt]
		\end{tabular}
		\label{tbl3}
		\begin{tablenotes}
			\footnotesize
			\item[*] \textbf{VGG16 will be removed during validation because it performs poorly during training.}
		\end{tablenotes}
	\end{threeparttable}
\end{table*}

\begin{figure}
	\centering
	\includegraphics[width = 1.0\linewidth]{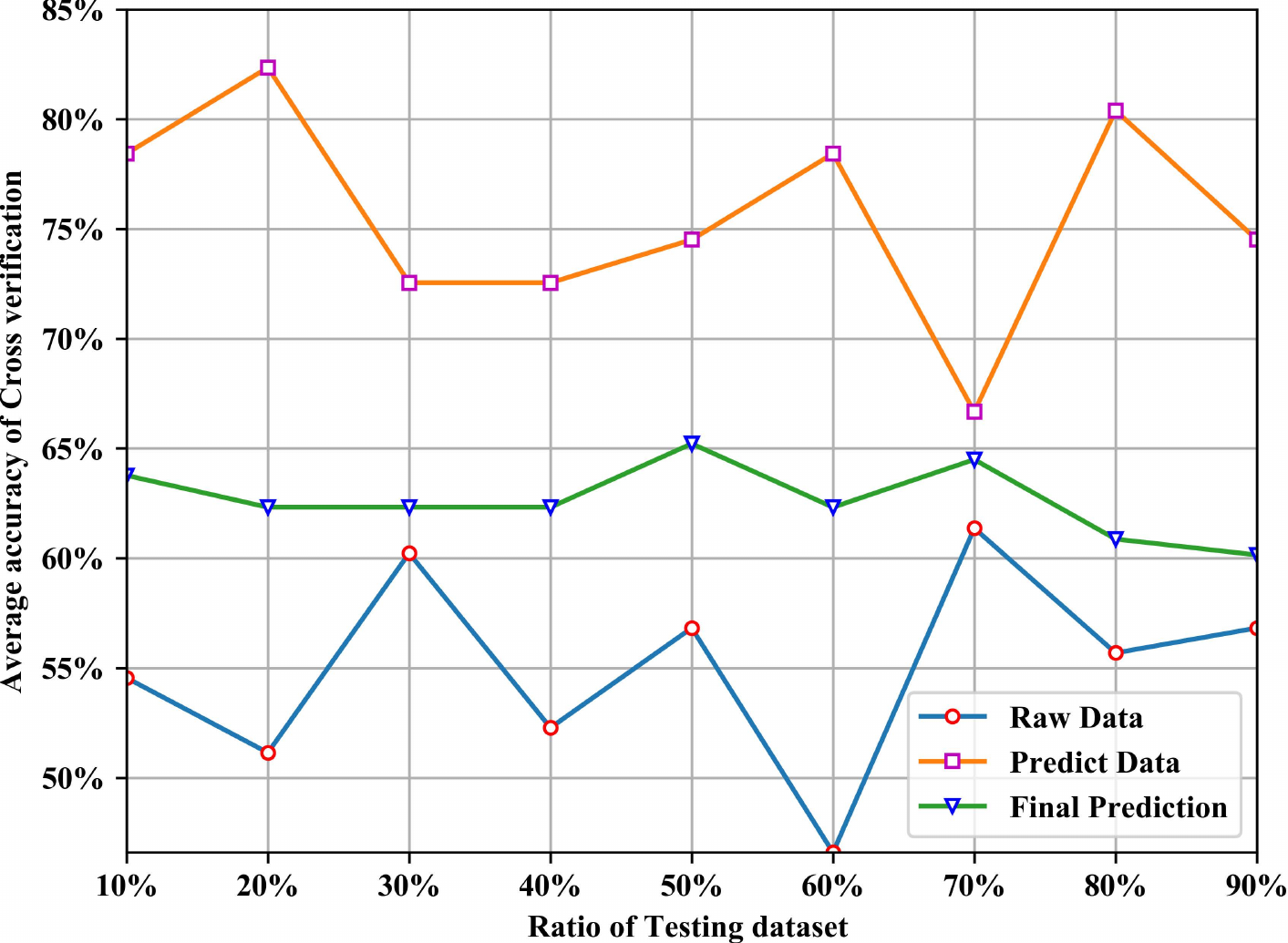}
	\caption{The comparison of the accuracy of the testing dataset divided by different proportions.}
	\label{fig:7}
\end{figure}

In order to further verify the learning ability of the model trained by generate data in the third part, we will train and verify the testing dataset which is also a double label testing dataset of patients and controls according to the proportion of pre-training, the type of dataset and the proportion of testing dataset. It is excluded that the proportion of training dataset is too small or too large, which leads to the distortion of model verification effect (i.e. the ratio is about 0.1 or 0.9). In the process of pre-training, we notice that the model after pre-training with generated data can achieve good results when the ratio is 0.6 (\textcolor{black!80}{Fig.~\protect\ref{fig:7}}, orange line). When the blank model used for control directly tests raw data without pre-training, the effect is rather unstable and the accuracy is relatively low (\textcolor{black!80}{Fig.~\protect\ref{fig:7}}, blue line). The discrimination ability of the final model, no matter what proportion of validation, will be higher than that of the untrained control model (\textcolor{black!80}{Fig.~\protect\ref{fig:7}}, green line).
\begin{figure}
	\centering
	\includegraphics[width = 1.0\linewidth]{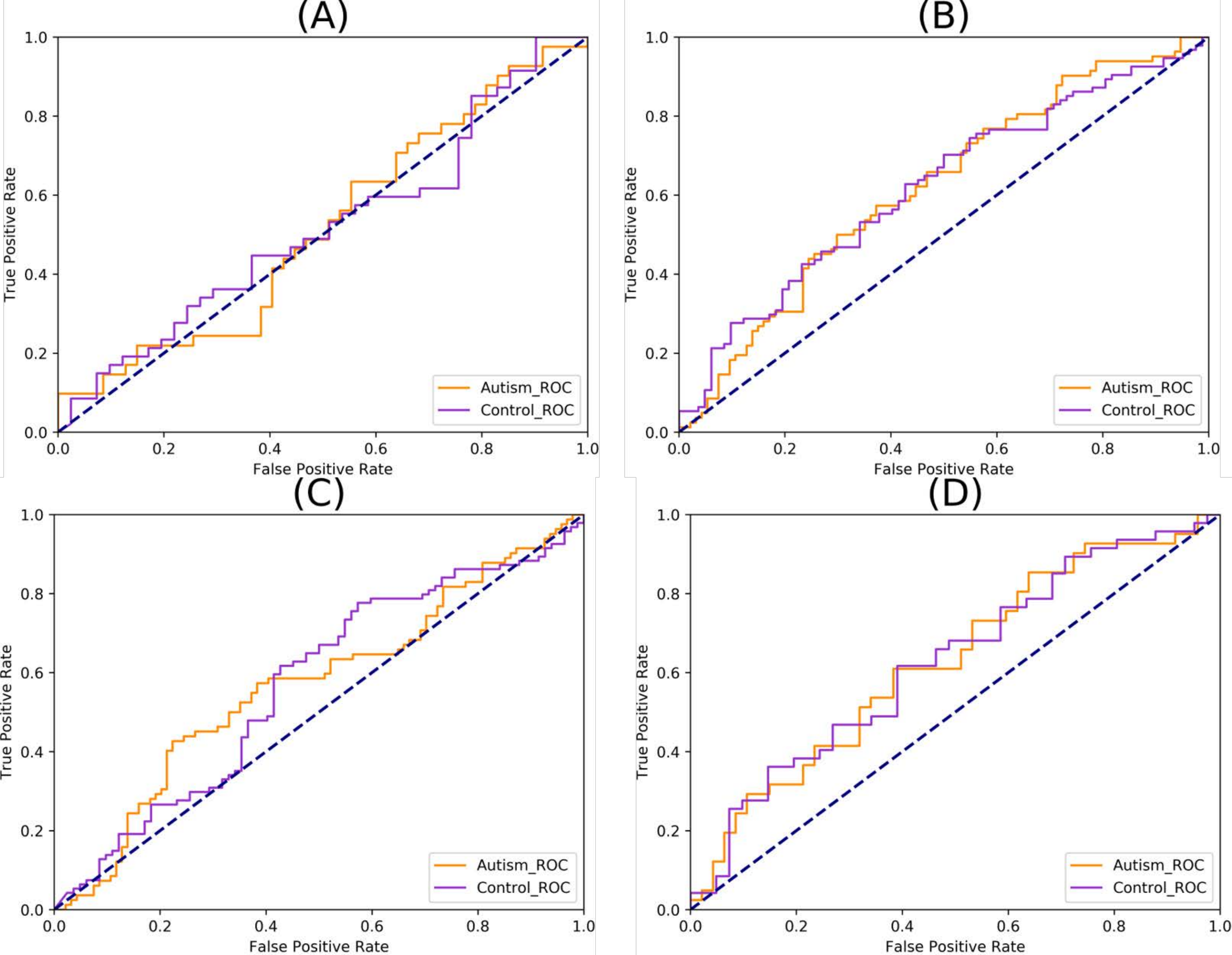}
	\caption{The ROC curves of patients and controls were predicted by generate data pre-training model. (A)Raw data training model to forecast raw data ratio=0.1; (B)Generate data training model to forecast raw data ratio=0.1; (C)Generate data training model to forecast raw data ratio=0.6; (D)Generate data training model to forecast raw data ratio=0.9.}
	\label{fig:8}
\end{figure}
Therefore, we choose the best model and proportion, i.e. pre-training Resnet-50 model with ratio of 0.6. We test the effect of the three datasets in \textcolor{black!80}{Table~\protect\ref{tbl3}} and adjust the test proportion to 0.1 for comparison (\textcolor{black!80}{Fig.~\protect\ref{fig:8}}). Because the model pre-trains the generated data, the prediction of the generated dataset is much higher than that of other groups.

\subsection{Comparative experiments of generating models with optimal parameters}

According to the experimental results in the previous section, we select the optimal hyper-parameters and strategies for the dataset and model. Based on these, this paper compares the traditional degree based and clustering based generation networks, as well as the comparative experiments using GraphRNN, GraphMLP and GraphVAE.

In the specific experiment, different generation models were used to learn the data of the original patients and the control group under the same threshold and other parameters under the optimal strategy of this experiment. For example, the GraphRNN after learning the real graph data, will get a model GraphRNN-A to generate the brain network of Autism, and a model GraphRNN-C for Control. These models are then used to generate a set of generated datasets with the same number of samples, which are mixed into the original dataset to train ResNet50 discriminant model. Finally, the ResNet50 model is used to test the original data in the same ratio to evaluate the learning effect.

\begin{figure}
	\centering
	\includegraphics[width = 1.0\linewidth]{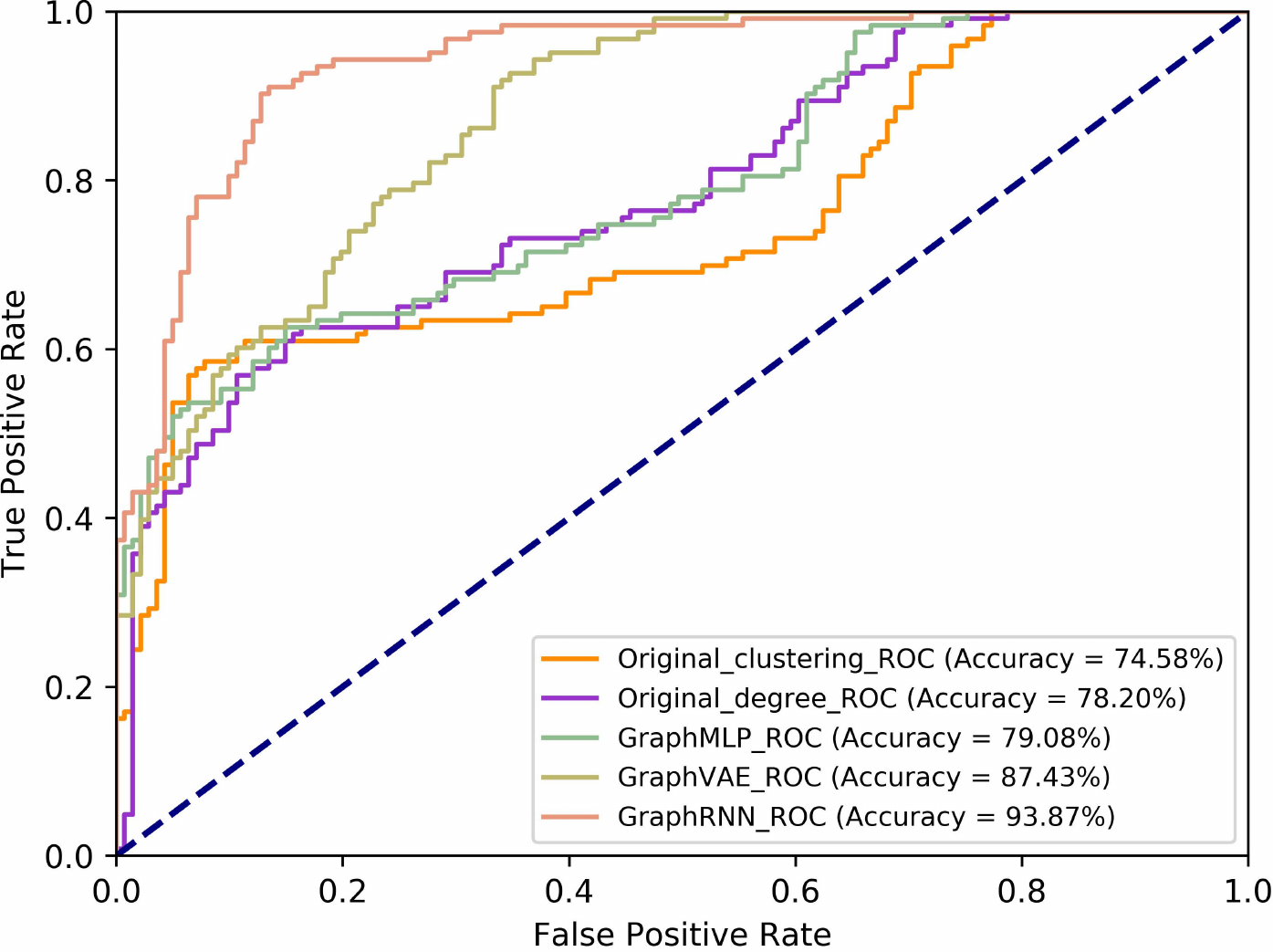}
	\caption{The ROC curves of control experiment was generated by different models with the same parameters in Autism.}
	\label{fig:9}
\end{figure}

After the cross validation of the original dataset not added to the generated data, we draw the ROC curves of the validation results of different generation models ((\textcolor{black!80}{Fig.~\protect\ref{fig:9}}).  It can be seen that due to the guidance and participation of the generated model, ResNet50 has significantly improved the accuracy (61.36$\%$) of identifying the original ABIDE dataset that is not added to the generated data compared with the original ResNet50 model (\textcolor{black!80}{Table~\protect\ref{tbl3}}) that is not reinforced by the generated data. The comparison results of different generation models show that GraphRNN is more prominent for the guidance model, which improves the accuracy of 32.51$\%$ compared with the previous ResNet50 model.

\section{Results and discussion}
Due to the high heterogeneity of Autism brain datasets, the difficulty of distinguishing brain network data features, the excessive number of graph edges, and the interference of other factors on the final generation and discrimination model, the accuracy of the data will decrease by 2.27$\%$ to 10.22$\%$ compared with the untreated original data, and the accuracy of some preprocessing methods will even differ by 16.84$\%$ under the same conditions. This study proves that the selection of preprocessing strategy in the early stage of brain formation model is particularly important for the effect of model discrimination.

The generated data is generated from the raw data, but in most of the training, using the same model, these data have a higher level of recognition, which is 16.24$\%$ to 25.02$\%$ higher than the raw data in different ResNet model. We speculate that in the model training, we constantly improve the probability that patients or controls may have edge connection, and set the unimportant factors to 0, resulting in the final generated graph features of patients and controls are more easily distinguished by the model.

In addition, it is noted that the reinforcement of generated data also improves the discrimination of ResNet model on the original ABIDE dataset. The most significant effect is the 50-layer ResNet, and the best generation model is GraphRNN, which improves the accuracy by 32.51$\%$ compared with the model reference experiment without generation data reinforcement. It might be that the data model distribution is learned in advance, which makes the model more inclined to extract the features with high edge connection probability in the generated model training \citep{RN65}, so it is also a good pre-training method in essence.

Although the pre-training of GraphRNN generation makes the model achieve good results, there is still no way to deal with the brain network formed at pixel level due to too many edges and nodes. In addition, the interaction ability of discriminator to generator is weak, and the principle of generating data depends on brain function connection matrix to a great extent. In the future work, we will further optimize the generating unit so that the model can accommodate more detailed graphs. And it will increase the guidance of discriminator to the generator, making the generator produce better brain network.

\section*{Acknowledge}
This work was supported by the Fundamental Research Funds for the Central Universities.(N2019006 and N180719020)  


\bibliographystyle{cas-model2-names}

\bibliography{cas-refs}


\end{document}